\documentclass[11pt]{article}
\usepackage{emnlp2016}

\makeatletter
\newcommand{\@BIBLABEL}{\@emptybiblabel}
\newcommand{\@emptybiblabel}[1]{}
\makeatother
\usepackage{hyperref}
\usepackage{url}

\usepackage{times}
\usepackage{latexsym}
\usepackage{amsmath}
\usepackage{multirow}
\usepackage{amsfonts}

\usepackage{textcomp}
\usepackage{color}
\usepackage{fancyvrb}
\usepackage{graphicx}
\usepackage{tabularx}
\usepackage{umoline}
\usepackage[linewidth=1pt]{mdframed}
\usepackage{environ}
\usepackage{tikz-dependency}

\NewEnviron{elaboration}{
\par
\begin{tikzpicture}
\node[rectangle,minimum width=\textwidth] (m) {\begin{minipage}{0.98\textwidth}\BODY\end{minipage}};
\draw[dashed] (m.south west) rectangle (m.north east);
\end{tikzpicture}
}

\newcommand{\mbf}[1]{\mathbf{#1}}
\newcommand{\fwd}[1]{\overrightarrow{#1}}
\newcommand{\bwd}[1]{\overleftarrow{#1}}
\newcommand{\enc}{\mathbf{enc}}
\newcommand{\gen}{\mathbf{gen}}
\newcommand{\x}{\mathbf{x}}
\newcommand{\y}{\mathbf{y}}
\newcommand{\z}{\mathbf{z}}
\newcommand{\h}{\mathbf{h}}
\newcommand{\loss}{\mathcal{L}}
\newcommand{\cost}{\text{cost}}

\definecolor{green}{rgb}{0.1,0.5,0.1}
\definecolor{red}{rgb}{1.0,0.2,0.2}

\newcommand{\add}[1]{{#1}}

\emnlpfinalcopy

\def\emnlppaperid{997}

\title{Rationalizing Neural Predictions}

\author{Tao Lei, Regina Barzilay and Tommi Jaakkola\\
Computer Science and Artificial Intelligence Laboratory\\
Massachusetts Institute of Technology\\
\texttt{\{taolei, regina, tommi\}@csail.mit.edu}}
\date{}

\begin{document}

\maketitle

\begin{abstract}
Prediction without justification has limited applicability. As a remedy, we learn to extract pieces of input text as justifications -- rationales -- that are tailored to be short and coherent, yet sufficient for making the same prediction. Our approach combines two modular components, generator and encoder, which are trained to operate well together. The generator specifies a distribution over text fragments as candidate rationales and these are passed through the encoder for prediction. Rationales are never given during training. Instead, the model is regularized by desiderata for rationales. We evaluate the approach on multi-aspect sentiment analysis against manually annotated test cases. Our approach outperforms attention-based baseline by a significant margin. We also successfully illustrate the method on the question retrieval task.\footnote{Our code and data are available at \url{https://github.com/taolei87/rcnn}.}
\end{abstract}
\section{Introduction}

Many recent advances in NLP problems have come from formulating and training expressive and elaborate neural models. This includes models for sentiment classification, parsing, and machine translation among many others. The gains in accuracy have, however, come at the cost of interpretability since complex neural models offer little transparency concerning their inner workings. In many applications, such as medicine,  predictions are used to drive critical decisions, including treatment options. It is necessary in such cases to be able to verify and understand the underlying basis for the decisions. Ideally, complex neural models would not only yield improved performance but would also offer interpretable justifications -- rationales -- for their predictions. 

\begin{figure}[t!]
%\begin{mdframed}
%\begin{elaboration}
%  \parbox{0.98\textwidth}{beer was n't what i expected , and i 'm not sure it 's `` true to style , '' but i thought it was delicious . \textbf{a very pleasant ruby red-amber color} with a relatively brilliant finish, but a limited amount of carbonation , from the look of it . aroma is what i think an amber ale should be - a nice blend of caramel and hoppiness bound together 
%}
%\end{elaboration}
%\vspace{-0.2cm}
%\begin{elaboration}
%\emph{Look}: 5 \\
%\emph{Smell}: 5  \\
%\emph{NumWounded}: 0 \\
%\emph{City}: Platte
%\end{elaboration}
%  \noindent\fbox{%  
%\end{mdframed}
\centering
\includegraphics[width=3.0in]{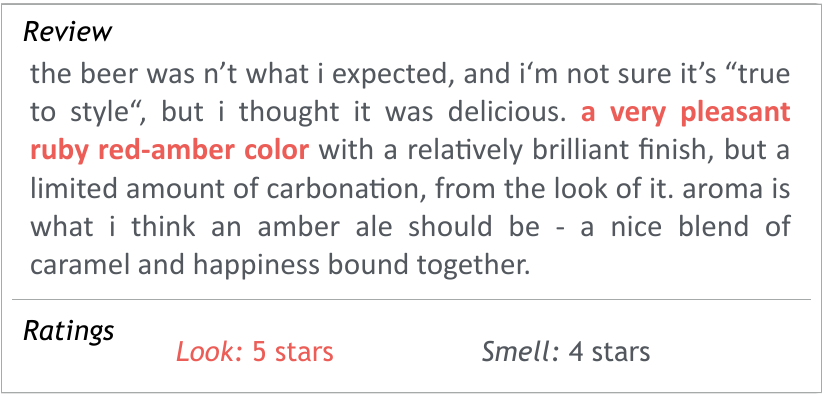}
\caption{An example of a review with ranking in two categories. The rationale for Look prediction is shown in bold.}
\label{Intro-Ex}
\end{figure}

In this paper, we propose a novel approach to incorporating rationale generation as an integral part of the overall learning problem. We limit ourselves to extractive (as opposed to abstractive) rationales. From this perspective, our rationales are simply subsets of the words from the input text that satisfy two key properties. First, the selected words represent short and coherent pieces of text (e.g., phrases) and, second, the selected words must alone suffice for prediction as a substitute of the original text. More concretely, consider the task of multi-aspect sentiment analysis. Figure~\ref{Intro-Ex} illustrates a product review along with user rating in terms of two categories or aspects. If the model in this case
predicts five star rating for color, it should also identify the phrase  \emph{"a very pleasant ruby red-amber color"} as the rationale underlying this decision. 

In most practical applications, rationale generation must be learned entirely in an unsupervised manner. We therefore assume that our model with rationales is trained on the same data as the original neural models, without access to additional rationale annotations. In other words, target rationales are never provided during training; the intermediate step of rationale generation is guided only by the two desiderata discussed above. Our model is composed of two modular components that we call the generator and the encoder. Our generator specifies a distribution over possible rationales (extracted text) and the encoder maps any such text to task specific target values. They are trained jointly to minimize a cost function that favors short, concise rationales while enforcing that the rationales alone suffice for accurate prediction.   

The notion of what counts as a rationale may be ambiguous in some contexts and the task of selecting rationales may therefore be challenging to evaluate. We focus on two domains where ambiguity is minimal (or can be minimized). The first scenario concerns with multi-aspect sentiment analysis exemplified by the beer review corpus~\cite{mcauley2012learning}. A smaller test set in this corpus identifies, for each aspect, the sentence(s) that relate to this aspect. We can therefore directly evaluate our predictions on the sentence level with the caveat that our model makes selections on a finer level, in terms of words, not complete sentences. The second scenario concerns with the problem of retrieving similar questions. 
The extracted rationales should capture the main purpose of the questions. We can therefore evaluate the quality of rationales as a compressed proxy for the full text in terms of retrieval performance.
%The titles in these questions can be taken as executive summaries of question bodies. Thus words selected from the title (if included as part of the body) can be assumed to represent reasonable rationales. In both cases, two versions of our model outperform predictions from simple linear classifiers or selection via attention based neural baseline. 
Our model achieves high performance on both tasks.
For instance, on the sentiment prediction task, our model achieves extraction accuracy of  96\%, as compared to 38\% and 81\% obtained by the bigram SVM and a neural attention baseline.

\section{Related Work}

Developing sparse interpretable models \add{is of considerable interest to} the broader research community\cite{LethamRuMcMa15,Kim_etal_nips15}. The need for interpretability is even more pronounced with recent neural models. Efforts in this area include analyzing and visualizing state activation~\cite{hermans2013training,karpathy2015visualizing,li:visualizing}, learning sparse interpretable word vectors~\cite{faruqui:2015:sparse}, and linking word vectors to semantic lexicons or word properties~\cite{faruqui:2015:Retro,herbelot-vecchi:2015:EMNLP}. 

\add{Beyond learning to understand or further constrain the network to be directly interpretable, one can estimate interpretable proxies that approximate the network. Examples include extracting ``if-then'' rules~\cite{thrun1995extracting} and decision trees~\cite{craven1996extracting} from trained networks. More recently, \newcite{ribeiro2016should} propose a model-agnostic framework where the proxy model is learned only for the target sample (and its neighborhood) thus ensuring locally valid approximations. Our work differs from these both in terms of what is meant by an explanation and how they are derived. In our case, an explanation consists of a concise yet sufficient portion of the text where the mechanism of selection is learned jointly with the predictor. 
}

Attention based models offer another means to explicate the inner workings of neural models~\cite{bahdanau2014neural,cheng2016long,MartinsA16,chen2015abc,xu2015ask,yang2015stacked}. Such models have been successfully applied to many NLP problems, improving both prediction accuracy as well as visualization and interpretability~\cite{rush2015neural,rocktaschel2015reasoning,hermann2015teaching}. \add{\newcite{icml2015_xuc15} introduced a stochastic attention mechanism together with a more standard soft attention on image captioning task. Our rationale extraction can be understood as a type of stochastic attention although architectures and objectives differ. Moreover, we compartmentalize rationale generation from downstream encoding so as to expose knobs to directly control types of rationales that are acceptable, and to facilitate broader modular use in other applications.}

%Attention based models offer another lens to the inner workings of neural models~\cite{bahdanau2014neural,cheng2016long,MartinsA16,chen2015abc,xu2015ask,yang2015stacked}. Such models have been successfully applied to many NLP problems, improving both prediction accuracy as well as visualization and interpretability~\cite{rush2015neural,rocktaschel2015reasoning,hermann2015teaching}.

%Our work differs from past approaches in terms of what is meant by interpretable models (generating concise yet sufficient rationales) and how interpretation is derived (rationale generation). We explicitly aim to identify salient portions of the input text to justify predictions. Moreover, architecturally, we detach the rationale generation from how it is used (encoding) so as to be able to directly control types of rationales that are acceptable, and to facilitate broader modular use in other applications. 

Finally, \add{we contrast our work with rationale-based classification~\cite{ZaidanEP07,marshall2015robotreviewer,ZhangMW16} which seek to improve prediction by relying on richer annotations in the form of human-provided rationales.} In our work, rationales are never given during training. The goal is to learn to generate them.

\section{Extractive Rationale Generation}

We formalize here the task of extractive rationale generation and illustrate it in the context of neural models. To this end, consider a typical NLP task where we are provided with a sequence of words as input, namely $\mbf{x}=\{x_1,\cdots, x_l\}$, where each $x_t\in\mathbb{R}^d$ denotes the vector representation of the i-th word. The learning problem is to map the input sequence $\mbf{x}$ to a target vector in $\mathbb{R}^m$. For example, in multi-aspect sentiment analysis each coordinate of the target vector represents the response or rating pertaining to the associated aspect. In text retrieval, on the other hand, the target vectors are used to induce similarity assessments between input sequences. Broadly speaking, we can solve the associated learning problem by estimating a complex parameterized mapping $\mbf{enc}(\mbf{x})$ from input sequences to target vectors. We call this mapping an \emph{encoder}. The training signal for these vectors is obtained either directly (e.g., multi-sentiment analysis) or via similarities (e.g., text retrieval). The challenge is that a complex neural encoder $\mbf{enc}(\mbf{x})$ reveals little about its internal workings and thus offers little in the way of justification for why a particular prediction was made. 

In extractive rationale generation, our goal is to select a subset of the input sequence as a \emph{rationale}. In order for the subset to qualify as a rationale it should satisfy two criteria: 1) the selected words should be interpretable and 2) they ought to suffice to reach nearly the same prediction (target vector) as the original input. In other words, a rationale must be short and sufficient. We will assume that a short selection is interpretable and focus on optimizing sufficiency under cardinality constraints.  

We encapsulate the selection of words as a \emph{rationale generator} which is another parameterized mapping $\mbf{gen}(\mbf{x})$ from input sequences to shorter sequences of words. Thus $\mbf{gen}(\mbf{x})$ must include only a few words and $\mbf{enc}(\mbf{gen}(\mbf{x}))$ should result in nearly the same target vector as the original input passed through the encoder or $\mbf{enc}(\mbf{x})$.  We can think of the generator as a tagging model where each word in the input receives a binary tag pertaining to whether it is selected to be included in the rationale.
%See Figure~\ref{fig:illustration}.
In our case, the generator is probabilistic and specifies a distribution over possible selections. 

The rationale generation task is entirely unsupervised in the sense that we assume no explicit annotations about which words should be included in the rationale. Put another way, the rationale is introduced as a latent variable, a constraint that guides how to interpret the input sequence. The encoder and generator are trained jointly, in an end-to-end fashion so as to function well together. 

\section{Encoder and Generator}

We use multi-aspect sentiment prediction as a guiding example to instantiate the two key components -- the encoder and the generator. The framework itself generalizes to other tasks. 

\paragraph{Encoder $\enc(\cdot)$:} Given a training instance $(\x,\y)$ where $\x=\{x_t\}_{t=1}^l$ is the input text sequence of length $l$ and $\y\in [0,1]^m$ is the target m-dimensional sentiment vector, the neural encoder predicts $\tilde{\y}=\enc(\x)$. If trained on its own, the encoder would aim to minimize the discrepancy between the predicted sentiment vector $\tilde{\y}$ and the gold target vector $\y$. We will use the squared error (i.e. $L_2$ distance) as the sentiment loss function,
\begin{align*}
\loss(\x,\y) = \|\tilde{\y}-\y\|_2^2 = \|\enc(\x)-\y\|_2^2
\end{align*}
The encoder could be realized in many ways such as a recurrent neural network. For example, let $\h_t = f_e(\x_t, \h_{t-1})$ denote a parameterized recurrent unit mapping input word $\x_t$ and previous state $\h_{t-1}$ to next state $\h_t$. The target vector is then generated on the basis of the final state reached by the recurrent unit after processing all the words in the input sequence. Specifically, 
\begin{align*}
\h_t &= f_e(\x_t, \h_{t-1}),\;\;t=1,\,\ldots, l\\
\tilde{\y} &= \sigma_e(\mbf{W}^e \h_l + \mbf{b}^e)
\end{align*}
%We will discuss these choices in more detail in Section~\ref{sec:implementation}.

\paragraph{Generator $\gen(\cdot)$:} The rationale generator extracts a subset of text from the original input $\x$ to function as an interpretable summary. Thus the rationale for a given sequence $\x$ can be equivalently defined in terms of binary variables $\{\z_1,\cdots,\z_l\}$ where each $\z_t\in {0,1}$ indicates whether word $\x_t$ is selected or not. From here on, we will use $\z$ to specify the binary selections and thus $(\z,\x)$ is the actual rationale generated (selections, input). We will use generator $\gen(\x)$ as synonymous with a probability distribution over binary selections, i.e., $\z\sim \gen(\x) \equiv p(\z|\x)$ where the length of $\z$ varies with the input $\x$. 

In a simple generator, the probability that the $t^{th}$ word is selected can be assumed to be conditionally independent from other selections given the input $\x$. That is, the joint probability $p(\z|\x)$ factors according to
\begin{align*}
p(\z|\x) &= \prod_{t=1}^l p(\z_t|\x) & (\text{independent selection})
\end{align*}
The component distributions $p(\z_t|\x)$ can be modeled using a shared bi-directional recurrent neural network. Specifically, let $\fwd{f}()$ and $\bwd{f}()$ be the forward and backward recurrent unit, respectively, then
\begin{align*}
\fwd{\h_t} &= \fwd{f}(\x_t, \fwd{\h_{t-1}}) \\
\bwd{\h_t} &= \bwd{f}(\x_t, \bwd{\h_{t+1}}) \\
p(\z_t|\x) &= \sigma_z(\mbf{W}^z [\fwd{\h_t};\bwd{\h_t}] + \mbf{b}^z)
\end{align*}
Independent but context dependent selection of words is often sufficient. However, the model is unable to select phrases or refrain from selecting the same word again if already chosen. To this end, we also introduce a dependent selection of words, 
\begin{align*}
p(\z|\x) = \prod_{t=1}^l p(\z_t|\x,\z_1\cdots\z_{t-1})
\end{align*}
which can be also expressed as a recurrent neural network. To this end, we introduce another hidden state $\mbf{s}_t$ whose role is to couple the selections. For example,
\begin{align*}
p(\z_t|\x,\z_{1,t-1}) &= \sigma_z(\mbf{W}^z [\fwd{\h_t};\bwd{\h_t};\mbf{s}_{t-1}] + \mbf{b}^z) \\
\mbf{s}_t &= f_z([\fwd{\h_t};\bwd{\h_t};\z_t], \mbf{s}_{t-1})
\end{align*}

\paragraph{Joint objective:} A rationale in our definition corresponds to the selected words, i.e., $\{\x_k|\z_k=1\}$. We will use $(\z,\x)$ as the shorthand for this rationale and, thus, $\enc(\z,\x)$ refers to the target vector obtained by applying the encoder to the rationale as the input. Our goal here is to formalize how the rationale can be made short and meaningful yet function well in conjunction with the encoder. Our generator and encoder are learned jointly to interact well but they are treated as independent units for modularity. 

The generator is guided in two ways during learning. First, the rationale that it produces must suffice as a replacement for the input text. In other words, the target vector (sentiment) arising from the rationale should be close to the gold sentiment. The corresponding loss function is given by
\begin{align*}
\loss(\z,\x,\y) = \|\enc(\z,\x)-\y\|_2^2
\end{align*}
Note that the loss function depends directly (parametrically) on the encoder but only indirectly on the generator via the sampled selection. 

Second, we must guide the generator to realize short and coherent rationales. It should select only a few words and those selections should form phrases (consecutive words) rather than represent isolated, disconnected words. We therefore introduce an additional regularizer over the selections
\begin{align*}
\Omega(\z) &= \lambda_1\|\z\| + \lambda_2 \sum_t |\z_t-\z_{t-1}|
\end{align*}
where the first term penalizes the number of selections while the second one discourages transitions (encourages continuity of selections). Note that this regularizer also depends on the generator only indirectly via the selected rationale. This is because it is easier to assess the rationale once produced rather than directly guide how it is obtained. 

Our final cost function is the combination of the two, $\cost(\z,\x,\y)=\loss(\z,\x,\y)+\Omega(\z)$. Since the selections are not provided during training, we minimize the expected cost: 
\begin{align*}
\min_{\theta_e,\theta_g} \sum_{(\x,\y)\in D}\mathbb{E}_{\z\sim \gen(\x)}\left[ \cost(\z,\x,\y) \right]
\end{align*}
where $\theta_e$ and $\theta_g$ denote the set of parameters of the encoder and generator, respectively, and $D$ is the collection of training instances. Our joint objective encourages the generator to compress the input text into coherent summaries that work well with the associated encoder it is trained with. 

Minimizing the expected cost is challenging since it involves summing over all the possible choices of rationales $\z$. This summation could potentially be made feasible with additional restrictive assumptions about the generator and encoder. However, we assume only that it is possible to efficiently sample from the generator. 

\paragraph{Doubly stochastic gradient} We now derive a sampled approximation to the gradient of the expected cost objective. This sampled approximation is obtained separately for each input text $\x$ so as to work well with an overall stochastic gradient method. Consider therefore a training pair $(\x,\y)$. For the parameters of the generator $\theta_g$,

\begin{small}
\vspace{-0.1in}
\begin{align*}
&\frac{\partial \mathbb{E}_{\z\sim \gen(\x)}\left[ \cost(\z,\x,\y) \right]}{\partial \theta_g} \\
&\quad=\quad\sum_{\z} \cost(\z,\x,\y) \cdot \frac{\partial p(\z|\x)}{\partial\theta_g} \\
&\quad=\quad\sum_{\z} \cost(\z,\x,\y) \cdot \frac{\partial p(\z|\x)}{\partial\theta_g} \cdot \frac{p(\z|\x)}{p(\z|\x)}
\end{align*}
\end{small}
Using the fact $(\log f(\theta))'= f'(\theta)/f(\theta)$, we get

\begin{small}
\vspace{-0.1in}
\begin{align*}
&\sum_{\z} \cost(\z,\x,\y) \cdot \frac{\partial p(\z|\x)}{\partial\theta_g} \cdot \frac{p(\z|\x)}{p(\z|\x)} \\
&\quad=\quad \sum_\z \cost(\z,\x,\y) \cdot \frac{\partial \log p(\z|\x)}{\partial \theta_g} \cdot p(\z|\x)\\
&\quad=\quad\mathbb{E}_{z\sim \gen(\x)}\left[ \cost(\z,\x,\y) \frac{\partial \log p(\z|\x)}{\partial \theta_g}\right]
\end{align*}
\end{small}
The last term is the expected gradient where the expectation is taken with respect to the generator distribution over rationales $\z$. Therefore, we can simply sample a few rationales $\z$ from the generator $\gen(\x)$ and use the resulting average gradient in an overall stochastic gradient method. A sampled approximation to the gradient with respect to the encoder parameters $\theta_e$ can be derived similarly,
\begin{align*}
&\frac{\partial \mathbb{E}_{\z\sim \gen(\x)}\left[ \cost(\z,\x,\y) \right]}{\partial \theta_e}\\
&\quad=\quad \sum_\z \frac{\partial\cost(\z,\x,\y)}{\partial\theta_e} \cdot p(\z|\x) \\
&\quad=\quad \mathbb{E}_{z\sim \gen(\x)}\left[ \frac{\partial\cost(\z,\x,\y)}{\partial\theta_e}\right]
\end{align*}

\paragraph{Choice of recurrent unit} We employ recurrent convolution (RCNN), a refinement of local-ngram based convolution. RCNN attempts to learn n-gram features that are not necessarily consecutive, and average features in a dynamic (recurrent) fashion. Specifically, for bigrams (filter width $n=2$) RCNN computes $\mbf{h}_t = f(\mbf{x}_t, \mbf{h}_{t-1})$ as follows

\begin{small}
\vspace{-0.1in}
\begin{align*}
    \mbf{\lambda}_t &= \sigma(\mbf{W}^{\lambda}\mbf{x}_t +
    \mbf{U}^\lambda\mbf{h}_{t-1}+\mbf{b}^\lambda) \\
    \mbf{c}^{(1)}_t &= \mbf{\lambda}_t\odot \mbf{c}^{(1)}_{t-1} +
    (1-\mbf{\lambda}_t)\odot (\mbf{W}_1\mbf{x}_t) \\
    \mbf{c}^{(2)}_t &= \mbf{\lambda}_t\odot \mbf{c}^{(2)}_{t-1} +
    (1-\mbf{\lambda}_t)\odot (\mbf{c}^{(1)}_{t-1}+\mbf{W}_2\mbf{x}_t) \\
    \mbf{h}_t &= \tanh(\mbf{c}^{(2)}_t + \mbf{b})
\end{align*}
\end{small}
RCNN has been shown to work remarkably in classification and retrieval applications~\cite{Lei15,lei2015qr} compared to other alternatives such CNNs and LSTMs. We use it for all the recurrent units introduced in our model.
\section{Experiments}

We evaluate the proposed joint model on two NLP applications: (1) multi-aspect sentiment analysis on product reviews and (2) similar text retrieval on AskUbuntu question answering forum.

\subsection{Multi-aspect Sentiment Analysis}

\begin{table}[t!]
\small
\centering
\begin{tabular}{l|l}
\hline
Number of reviews & 1580k \\
Avg length of review & 144.9 \\
Avg correlation between aspects  & 63.5\% \\
Max correlation between two aspects & 79.1\% \\
Number of annotated reviews & 994 \\
\hline
\end{tabular}
\caption{Statistics of the beer review dataset.}
\label{table:beerdataset}
\end{table}

\begin{table*}[t!]
\small
\centering
\begin{tabular}{l|cc|cc|cc}
\hline
\multirow{2}{*}{Method} & \multicolumn{2}{c|}{Appearance} & \multicolumn{2}{c|}{Smell} & \multicolumn{2}{c}{Palate} \\
\cline{2-7}
& \% precision & \% selected & \% precision & \% selected & \% precision & \% selected  \\
\hline
SVM &  38.3 & 13 & 21.6 & 7 & 24.9 & 7 \\
%\hline
Attention model & 80.6 & 13 & 88.4 & 7 & 65.3 & 7 \\
\hline
Generator (independent) & 94.8 & 13 & 93.8 & 7 & 79.3 & 7 \\
%& 97.4 & 6 & 94.6 & 5 & 89.8 & 4 \\
%\hline
Generator (recurrent) & 96.3 & 14 & 95.1 & 7 & 80.2 & 7 \\
%& - & - & - & - & - & - \\
\hline
\end{tabular}
\caption{Precision of selected rationales for the first three aspects. The
precision is evaluated based on whether the selected words are in the sentences
describing the target aspect, based on the sentence-level annotations. Best training epochs are selected based on the objective value on the development set (no sentence annotation is used).}
\label{table:beerprecision}
\end{table*}

\paragraph{Dataset} We use the BeerAdvocate\footnote{\url{www.beeradvocate.com}}
review dataset used in prior work~\cite{mcauley2012learning}.\footnote{\url{http://snap.stanford.edu/data/web-BeerAdvocate.html}} This dataset contains 1.5 million reviews written by the website users. The reviews are naturally multi-aspect -- each of them contains multiple sentences describing the \emph{overall} impression or one particular aspect of a beer, including \emph{appearance}, \emph{smell} (aroma), \emph{palate} and the \emph{taste}. In addition to the
written text, the reviewer provides the ratings (on a scale of 0 to 5 stars) for each aspect as well as an overall rating. The ratings can be fractional (e.g. 3.5 stars), so we normalize the scores to $[0,1]$ and use them as the (only) supervision for regression.

\newcite{mcauley2012learning} also provided sentence-level annotations on around 1,000 reviews. Each sentence is annotated with one (or multiple) aspect label, indicating what aspect this sentence covers. We use this set as our test set to evaluate the precision of words in the extracted rationales.

\begin{table}[t!]
\small
\centering
\begin{tabular}{l|cccc|c}
\hline
 & $D$ & $d$ & $l$ & $\vert\theta\vert$ & MSE \\
\hline
SVM & 260k & - & - & 2.5M & 0.0154 \\
SVM & 1580k & - & - & 7.3M & 0.0100 \\
\hline
LSTM & 260k & 200 & 2 & 644k & 0.0094 \\
RCNN & 260k & 200 & 2 & 323k & \textbf{0.0087} \\
\hline
\end{tabular}
\caption{Comparing neural encoders with bigram SVM model. MSE is the mean
squared error on the test set. $D$ is the amount of data used for training and
development. $d$ stands for the hidden dimension, $l$
denotes the depth of network and $\vert\theta\vert$ denotes the number of parameters (number of features for SVM).}
\label{table:encoder}
\end{table}

Table~\ref{table:beerdataset} shows several statistics of the beer review
dataset. The sentiment correlation between any pair of aspects (and the overall score) is quite high, getting 63.5\% on average and a maximum of 79.1\% (between the \emph{taste} and \emph{overall} score). If directly training the model on this set, the model can be confused due to such strong correlation. We therefore perform a preprocessing step, picking ``less correlated'' examples from the dataset.\footnote{Specifically, for each aspect we train a simple linear regression model to predict the rating of this aspect given the ratings of the other four aspects. We then keep picking reviews with largest prediction error
until the sentiment correlation in the selected subset increases dramatically.}
This gives us a de-correlated subset for each aspect, each containing
about 80k to 90k reviews. We use 10k as the development set. We focus on three aspects since the fourth aspect \emph{taste} still gets $>50\%$ correlation with the overall sentiment.
%After this process, the appearance, smell and palate
%aspects achieve a maximum of XX, YY and ZZ correlation with others. The taste
%aspect still gets about 50\% correlation with the overall sentiment.

\begin{figure}[t!]
\vspace{-0.03in}
\centering
\includegraphics[width=2.9in]{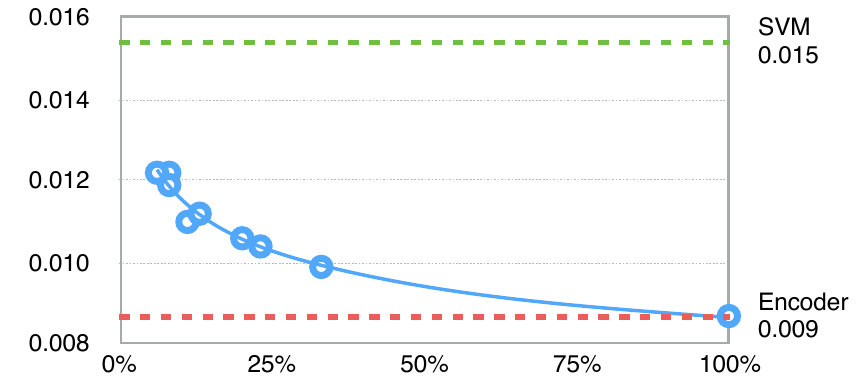}
\vspace{-0.08in}
\caption{Mean squared error of all aspects on the test set (y-axis) when various percentages of text are extracted as rationales (x-axis). 220k training data is used.}
\label{figure:tradeoff}
\end{figure}

\paragraph{Sentiment Prediction} Before training the joint model, it is worth assessing the neural encoder separately to check how accurately the neural network
predicts the sentiment. To this end, we compare neural encoders with bigram SVM model, training medium and large SVM models using 260k and all 1580k reviews respectively. As shown in Table~\ref{table:encoder}, the recurrent neural network models outperform the SVM model for sentiment prediction and also require less training data to achieve the performance. The LSTM and RCNN units obtain similar test error, getting 0.0094 and 0.0087 mean squared error respectively. The RCNN unit performs slightly better and uses less parameters. Based on the results, we choose the RCNN encoder network with 2 stacking layers and 200 hidden states. 

To train the joint model, we also use RCNN unit with 200 states as the forward and backward recurrent unit for the generator $\gen()$. The dependent generator has one additional recurrent layer. For this layer we use 30 states so the dependent version still has a number of parameters comparable to the independent version. The two versions of the generator have 358k and 323k parameters respectively.

Figure~\ref{figure:tradeoff} shows the performance of our joint dependent model when trained to predict the sentiment of all aspects. We vary the regularization $\lambda_1$ and $\lambda_2$ to show various runs that extract different amount of text as rationales. Our joint model gets performance close to the best encoder run (with full text) when few words are extracted.

\begin{figure*}[t!]
\vspace{-0.06in}
\centering
\includegraphics[width=6.35in]{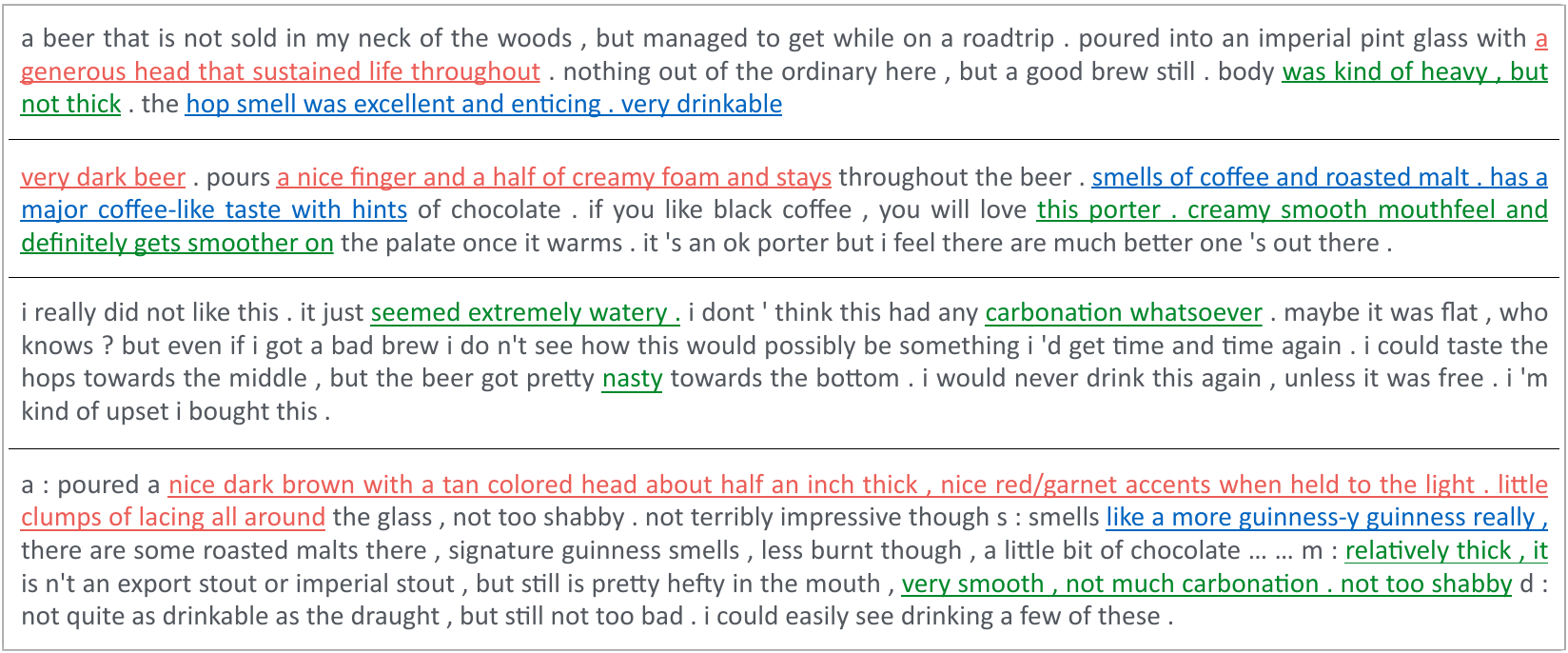}
\vspace{-0.05in}
\caption{Examples of extracted rationales indicating the sentiments of various aspects. The extracted texts for appearance, smell and palate are shown in red, blue and green color respectively. The last example is shortened for space.}
\label{figure:review_rationales}
\end{figure*}

\begin{figure}[t!]
\vspace{-0.04in}
\centering
\includegraphics[width=2.4in]{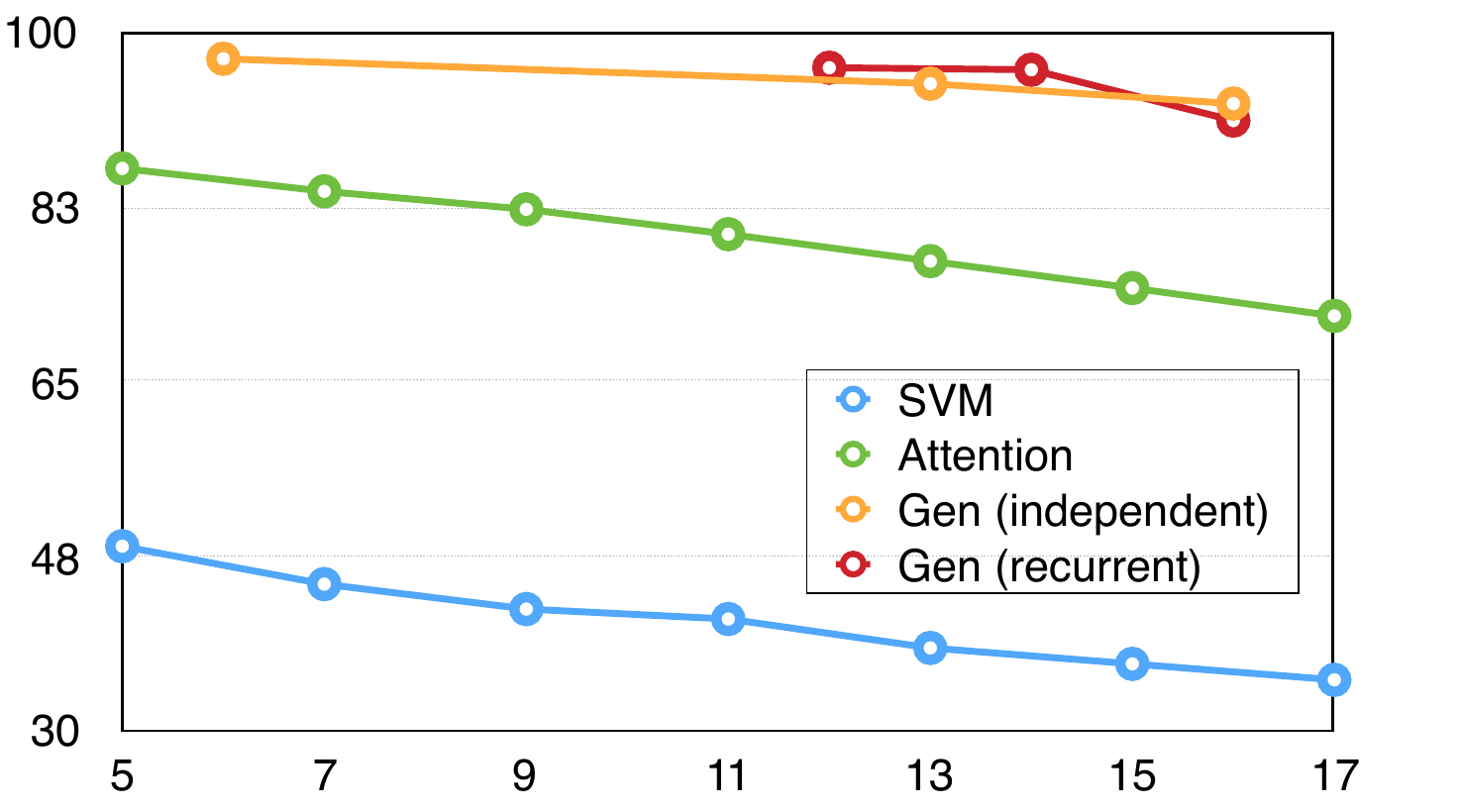}
\vspace{-0.05in}
\caption{Precision (y-axis) when various percentages of text are extracted as rationales (x-axis) for the appearance aspect.}
\label{figure:precisions}
\vspace{-0.1in}
\end{figure}

\paragraph{Rationale Selection} To evaluate the supporting rationales for each aspect, we train the joint encoder-generator model on each de-correlated subset. 
We set the cardinality regularization $\lambda_1$ between values $\{2e-4,3e-4,4e-4\}$ so the extracted rationale texts are neither too long nor too short. For simplicity, we set $\lambda_2=2\lambda_1$ to encourage local coherency of the extraction.

For comparison we use the bigram SVM model and implement an attention-based neural network model. The SVM model successively extracts unigram or bigram (from the test reviews) with the highest feature. The attention-based model learns a normalized attention vector of the input tokens (using similarly the forward and backward RNNs), then the model averages over the encoder states accordingly to the attention, and feed the averaged vector to the output layer. Similar to the SVM model, the attention-based model can selects words based on their attention weights. 

\begin{figure}[t!]
\vspace{-0.1in}
\centering
\includegraphics[width=2.6in]{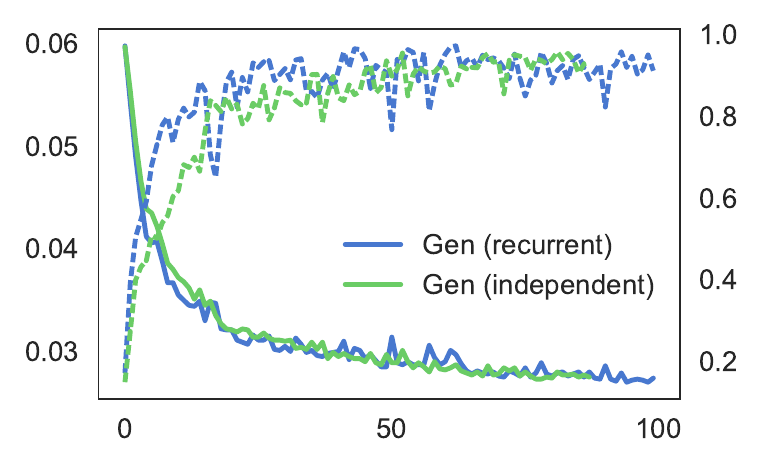}
\vspace{-0.12in}
\caption{Learning curves of the optimized cost function on the development set and the precision of rationales on the test set. The smell (aroma) aspect is the target aspect.}
\label{figure:learningcurves}
\vspace{-0.11in}
\end{figure}

Table~\ref{table:beerprecision} presents the precision of the extracted rationales calculated based on sentence-level aspect annotations. The $\lambda_1$ regularization hyper-parameter is tuned so the two versions of our model extract similar number of words as rationales. The SVM and attention-based model are constrained similarly for comparison.   Figure~\ref{figure:precisions} further shows the precision when different amounts of text are extracted. Again, for our model this corresponds to changing the $\lambda_1$ regularization.
As shown in the table and the figure, our encoder-generator networks extract text pieces describing the target aspect with high precision, ranging from 80\% to 96\% across the three aspects appearance, smell and palate. The SVM baseline performs poorly, achieving around 30\% accuracy.
The attention-based model achieves reasonable but worse performance than the rationale generator, suggesting the potential of directly modeling rationales as explicit extraction.

Figure~\ref{figure:learningcurves} shows the learning curves of our model for the smell aspect. In the early training epochs, both the independent and (recurrent) dependent selection models fail to produce good rationales, getting low precision as a result. After a few epochs of exploration however, the models start to achieve high accuracy. We observe that the dependent version learns more quickly in general, but both versions obtain close results in the end.

Finally we conduct a qualitative case study on the extracted rationales. Figure~\ref{figure:review_rationales} presents several reviews, with highlighted rationales predicted by the model. Our rationale generator identifies key phrases or adjectives that indicate the sentiment of a particular aspect.

\subsection{Similar Text Retrieval on QA Forum}

\paragraph{Dataset} For our second application, we use the real-world AskUbuntu\footnote{\url{askubuntu.com}} dataset used in recent work~\cite{dossantos-EtAl:2015,lei2015qr}. This set contains a set of 167k unique questions (each consisting a question title and a body) and 16k user-identified similar question pairs. Following previous work, this data is used to train the neural encoder that learns the vector representation of the input question, optimizing the cosine distance (i.e. cosine similarity) between similar questions against random non-similar ones. We use the ``one-versus-all'' hinge loss (i.e. positive versus other negatives) for the encoder, similar to~\cite{lei2015qr}. During development and testing, the model is used to score 20 candidate questions given each query question, and a total of $400\times 20$ query-candidate question pairs are annotated for evaluation\footnote{\url{https://github.com/taolei87/askubuntu}}.

\paragraph{Task/Evaluation Setup} The question descriptions are often long and fraught with irrelevant details. 
In this set-up, a fraction of the original question text should be sufficient to represent its content, and be used for retrieving similar questions.  Therefore, we will evaluate rationales based on the accuracy of the question retrieval task, assuming that better rationales achieve higher performance. To put this performance in context, we also report 
the accuracy when full body of a question is used, as well as titles alone. The latter constitutes an upper bound on the model performance as in this dataset titles provide short, informative summaries of the question content. 
We evaluate the rationales using the mean average precision (MAP) of retrieval.
% describe the training objective for this task ?

\begin{table}[t!]
%\vspace{-0.06in}
\footnotesize
\centering
\begin{tabular}{l|ccc}
\hline
 & MAP (dev) & MAP (test) & \%words \\
\hline
Full title & 56.5 & 60.0 & 10.1 \\
Full body & 54.2 & 53.0 & 89.9 \\
\hline
 \multirow{2}{*}{Independent} & 55.7 & 53.6 & 9.7\\
 & 56.3 & 52.6 & 19.7 \\
 \hline
\multirow{2}{*}{Dependent} & 56.1 & 54.6 & 11.6 \\
& 56.5 & 55.6 & 32.8 \\
 \hline
\end{tabular}
\caption{Comparison between rationale models (middle and bottom rows) and the baselines using full title or body (top row).}
\label{table:qr_overall}
\end{table}

\begin{figure}[t!]
%\vspace{-0.06in}
\centering
\includegraphics[width=2.2in]{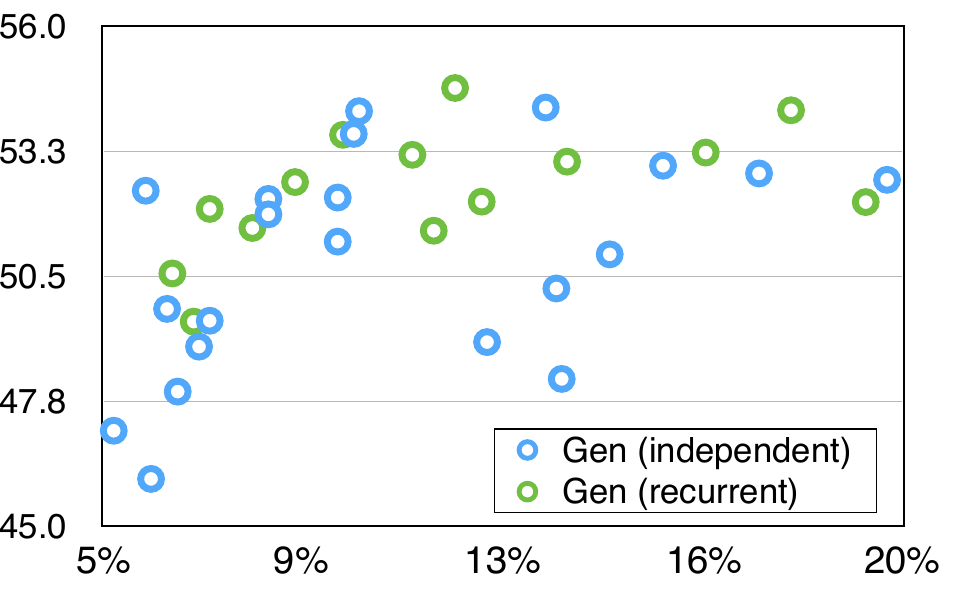}
%\vspace{-0.09in}
\caption{Retrieval MAP on the test set when various percentages of the texts are chosen as rationales. Data points correspond to models trained with different hyper-parameters.}
\label{figure:qr_map}
\end{figure}

\begin{figure*}[t!]
%\vspace{-0.06in}
\centering
\includegraphics[width=6.45in]{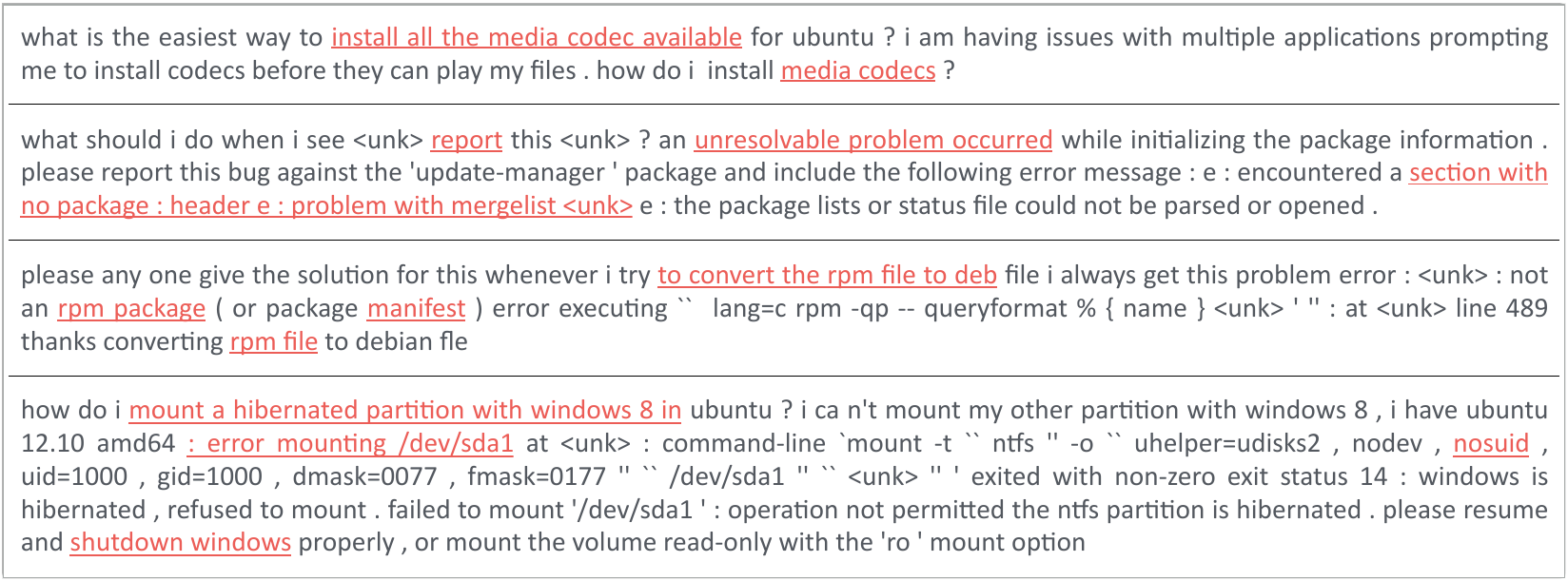}
%\vspace{-0.08in}
\caption{Examples of extracted rationales of questions in the AskUbuntu domain.}
\label{figure:question_rationales}
\end{figure*}

\paragraph{Results} Table~\ref{table:qr_overall} presents the results of our rationale model. We explore a range of hyper-parameter values\footnote{$\lambda_1\in\{.008, .01, .012, .015\}$, $\lambda_2=\{0,\lambda_1, 2\lambda_1\}$, dropout $\in \{0.1, 0.2\}$}. We include two runs for each version. The first one achieves the highest MAP on the development set, The second run is selected to compare the models when they use roughly 10\% of question text (7 words on average).
%two representative runs for the independent selection and recurrent selection model versions -- the run that selects roughly the amount of texts similar to the amount of title words (11$\sim$12\%) and (2) the run that achieves the best MAP on the dev set. 
We also show the results of different runs in Figure~\ref{figure:qr_map}. The rationales achieve the MAP up to 56.5\%, getting close to using the titles. The models also outperform the baseline of using the noisy question bodies, indicating the the models' capacity of extracting short but important fragments. 

Figure~\ref{figure:question_rationales} shows the rationales for several questions in the AskUbuntu domain, using the recurrent version with around 10\% extraction. Interestingly, the model does not always select words from the question title. The reasons are that the question body can contain the same or even complementary information useful for retrieval. Indeed, some rationale fragments shown in the figure are error messages, which are typically not in the titles but very useful to identify similar questions.
%Interestingly, the run that achieves the best MAP (i.e. 56.5/55.6 on development and test sets) is not the run that selects most words from the title text (i.e. 61.7). The reasons are that the body text can contain the same or even complementary information useful for retrieval. Indeed, , and some text pieces such as error messages (typically not in the titles) are intuitively important.

\section{Discussion}
\add{
We proposed a novel modular neural framework to automatically generate concise yet sufficient text fragments to justify predictions made by neural networks. We demonstrated that our encoder-generator framework, trained in an end-to-end manner, gives rise to quality rationales in the absence of any explicit rationale annotations. The approach could be modified or extended in various ways to other applications or types of data.

\paragraph{Choices of $\mathbf{enc(\cdot)}$ and $\mathbf{gen(\cdot)}$.} The encoder and generator can be realized in numerous ways without changing the broader algorithm. For instance, we could use a convolutional network~\cite{Kim14,kalchbrenner2014}, deep averaging network~\cite{iyyer2015,joulin2016bag} or a boosting classifier as the encoder. When rationales can be expected to conform to repeated stereotypical patterns in the text, a simpler encoder consistent with this bias can work better. We emphasize that, in this paper, rationales are flexible explanations that may vary substantially from instance to another. On the generator side, many additional constraints could be imposed to further guide acceptable rationales. 

\paragraph{Dealing with Search Space.} Our training method employs a REINFORCE-style algorithm~\cite{williams1992simple} where the gradient with respect to the parameters is estimated by sampling possible rationales. Additional constraints on the generator output can be helpful in alleviating problems of exploring potentially a large space of possible rationales in terms of their interaction with the encoder. We could also apply variance reduction techniques to increase stability of stochastic training~(cf. \cite{weaver2001optimal,mnih2014recurrent,ba-attention-2015,icml2015_xuc15}).
}
\section{Acknowledgments}
We thank Prof. Julian McAuley for sharing the review dataset and annotations. We also thank MIT NLP group and the reviewers for their helpful comments. \add{The work is supported by the Arabic Language Technologies (ALT) group at Qatar Computing Research Institute (QCRI) within the IYAS project.} Any opinions, findings, conclusions, or recommendations expressed in this paper are those of the authors, and do not necessarily reflect the views of the funding organizations.

\bibliography{paper}
\bibliographystyle{emnlp2016}

\end{document}